\documentclass[conference]{IEEEtran}
\IEEEoverridecommandlockouts
\usepackage{cite}
\usepackage{amsmath,amssymb,amsfonts}
\usepackage{algorithmic}
\usepackage{graphicx}
\usepackage{textcomp}
\usepackage{xcolor}
\def\BibTeX{{\rm B\kern-.05em{\sc i\kern-.025em b}\kern-.08em
    T\kern-.1667em\lower.7ex\hbox{E}\kern-.125emX}}

\usepackage{cite}
\usepackage{amsmath,amssymb,amsfonts}
\usepackage{algorithmic}
\usepackage{graphicx}
\usepackage{textcomp}
\usepackage{xcolor}

\usepackage{microtype}
\usepackage{subfigure}
\usepackage{booktabs} 
\usepackage{amsmath}
\usepackage{amssymb}
\usepackage{mathtools}
\usepackage{amsthm}
\usepackage{enumitem}
\usepackage{multirow}
\usepackage{makecell}

\usepackage{caption}

\usepackage{makecell}
\usepackage{bbding}
\usepackage{tabularx}
\usepackage{hyperref}

\usepackage{tikz}

\begin{document}

\title{Rapid-INR: Storage Efficient CPU-free DNN Training Using Implicit Neural Representation}

\author{\IEEEauthorblockN{Hanqiu Chen,
Hang Yang$^\S$, Stephen Fitzmeyer$^\S$, Cong Hao}

\thanks{\scriptsize $^{\S}$ Equal contribution.}

\IEEEauthorblockA{School of Electrical and Computer Engineering, Georgia Institute of Technology}
\IEEEauthorblockA{\{hanqiu.chen, hyang628, sfitzmeyer3, callie.hao\}@gatech.edu}
}

\maketitle

\begin{abstract}

Implicit Neural Representation (INR) is an innovative approach for representing complex shapes or objects without explicitly defining their geometry or surface structure. Instead, INR represents objects as continuous functions. Previous research has demonstrated the effectiveness of using neural networks as INR for image compression, showcasing comparable performance to traditional methods such as JPEG. However, INR holds potential for various applications beyond image compression. This paper introduces Rapid-INR, a novel approach that utilizes INR for encoding and compressing images, thereby accelerating neural network training in computer vision tasks. Our methodology involves storing the whole dataset directly in INR format on a GPU, mitigating the significant data communication overhead between the CPU and GPU during training. Additionally, the decoding process from INR to RGB format is highly parallelized and executed on-the-fly. To further enhance compression, we propose iterative and dynamic pruning, as well as layer-wise quantization, building upon previous work. We evaluate our framework on the image classification task, utilizing the ResNet-18 backbone network and three commonly used datasets with varying image sizes. Rapid-INR reduces memory consumption to only 5\% of the original dataset size and achieves a maximum 6$\times$ speedup over the PyTorch training pipeline, as well as a maximum 1.2$\times$ speedup over the DALI training pipeline, with only a marginal decrease in accuracy. Importantly, Rapid-INR can be readily applied to other computer vision tasks and backbone networks with reasonable engineering efforts.  Our implementation code is publicly available at \textcolor{blue}{\url{https://github.com/sharc-lab/Rapid-INR}}.

\end{abstract}


\section{Introduction}


In recent years, Deep Neural Networks (DNNs) have gained significant attention for their effectiveness in various AI tasks, such as computer vision~\cite{he2015deep,girshick2015fast}, natural language processing~\cite{sun2021interpreting,dinesh2021review}, healthcare~\cite{chen2021synthetic}, and autonomous driving~\cite{Grigorescu_2020,liu2022understanding}. The growing demand for on-device machine learning model training is driven by several factors, including the need to maintain data privacy, enable personalized models and lifelong learning, and enhance energy efficiency by reducing reliance on cloud-based data transmission.


However, as neural networks and datasets continue to grow in size and complexity, there are significant challenges in directly offloading the end-to-end training of deep neural networks (DNNs) onto a single device. \underline{\textbf{First}}, effectively reducing storage requirements for large training datasets while preserving essential information for training the neural networks is difficult. This is crucial to avoid the need for external memory data communication, which can slow down training and introduce latency. \underline{\textbf{Second}}, developing on-the-fly image decoding techniques is challenging, as it requires real-time and parallel decoding of compressed images during the training process.


To address these challenges, researchers are focusing on developing advanced image compression techniques that strike a balance between reducing storage requirements and preserving important training information. Implicit Neural Representation (INR) has emerged as a promising method in computer vision for reducing image size. However, existing studies~\cite{ramirez2022_0,strumpler2022implicit,zhang2021implicit} are mainly focusing on optimizing INR for high compression rates and minimal quality loss, overlooking its broader applications. Therefore, it is essential to thoroughly explore and leverage the full potential of INR to maximize its benefits.

Motivated by the considerable overhead associated with external memory access in computer vision tasks, as well as the impressive image compression capabilities of INR, this paper proposes 
an innovative framework \textbf{Rapid-INR}. 
The main purpose of this framework is to overcome the data transmission bottleneck by utilizing INR to compress the entire image dataset in multilayer perceptron (MLP) weights and achieve end-to-end training only on GPU. This compression enables on-device storage without the need for external memory access, while also facilitating on-the-fly decoding. The contributions of this paper can be summarized as follows:

\begin{itemize}[leftmargin=*]

\item \textbf{CPU-free training with exceptional speedup.} 
In contrast to conventional approaches that rely on powerful host devices (such as CPUs) and multiple data-loader workers to expedite image data pre-processing and transmission, Rapid-INR offers the advantage of offloading the entire training process onto GPU without the need for CPU and external memory accesses. This benefit stems from two key factors. Firstly, the INR format results in smaller image sizes. Secondly, images are stored in INR weight tensors, which are more compatible with GPU CUDA memory as compared to the JPEG format. This approach effectively reduces the data communication overhead and significantly improves training speed.

\item \textbf{High decoding parallelism without specialized hardware. 
} 
In the backbone training process, on-the-fly decoding of a batch of images from the INR weight format to RGB format is made possible by leveraging pixel-level parallelism, enabling simultaneous processing of individual pixels. By fully utilizing the CUDA cores in the INR decoding stage, we achieve optimal decoding speed without the need for specialized hardware.



\item \textbf{Optimized compression method for efficient storage.}
To further reduce memory consumption in existing INR-based image compression methods~\cite{dupont2021coin,dupont2022coin} while maintaining image quality, we introduce dynamic pruning and layer-wise quantization. Dynamic pruning selectively prunes INR weights based on reconstructed image quality, ensuring efficient memory usage. Layer-wise quantization involves quantizing weights to 8 bits in layers with minimal impact on final performance. By leveraging these techniques, we achieve significant memory savings while preserving image quality within acceptable limits.

\item \textbf{Ease of use with high generality.} One of the key advantages of Rapid-INR is its seamless integration with mainstream computer vision tasks training pipelines, requiring only minor modifications. Additionally, Rapid-INR is compatible with common data augmentation techniques. Unlike traditional approaches, Rapid-INR decouples the data representation from the spatial resolution. The memory needed to parameterize the signal is no longer dependent on the spatial resolution but rather scales with the complexity of the underlying signal. This enables INR-encoded images to be decoded to arbitrary spatial resolutions, offering greater flexibility and adaptability in downstream tasks.


\item \textbf{Neural architecture search.}
We conduct a comprehensive study by exploring different numbers of layers and hidden dimensions in the MLP used for INR to investigate the relationship between MLP architecture and reconstructed image quality. Our goal is to optimize the MLP architecture when the size of MLP used for INR is fixed while maintaining high-quality reconstructions.

\item  \textbf{Experiment results.} Rapid-INR outperforms existing frameworks with significant performance improvements. It achieves a maximum speedup of $6\times$ compared to the PyTorch training pipeline with one data-loader worker. Additionally, Rapid-INR achieves a maximum speedup of $1.2\times$ compared to the NVIDIA DALI framework with one CPU thread. Moreover, it maintains a high level of accuracy in image classification tasks, with only about a 2\% accuracy drop, while utilizing just about 5\% of the original JPEG storage space.

\end{itemize}

\section{Background and motivation}

\subsection{JPEG and Implicit Neural Representation}



JPEG (Joint Photographic Experts Group) is a popular image compression standard renowned for reducing image file sizes while maintaining visual quality. It achieves compression by employing mathematical transformations such as color space transformation, discrete cosine transform, and quantization. The quantization matrix used during the quantization stage is crucial in determining the size and quality of JPEG images. The choice of quality settings directly impacts the selection of the quantization matrix. Additionally, downsampling is an alternative method for JPEG compression.




Implicit Neural Representations (INR) present a novel approach to representing a wide range of signals~\cite{mildenhall2021nerf,park2019deepsdf,mescheder2019occupancy,chan2021pi,zhang2021implicit}. They employ a continuous function that maps the signal's domain (e.g., pixel coordinates in an image) to its corresponding value (e.g., the R, G, B color of that pixel). The goal of this study is to compress images that can be expressed using a set of coordinates $x \in X$ and RGB values $y \in Y$. Each data point consists of a collection of coordinate and RGB value pairs $d = \{(x_i, y_i)\}_{i=1}^{n}$. The objective is to train a neural network $f_\theta : X \rightarrow Y$ with parameters $\theta$  by minimizing the following loss function:

\begin{equation}
L(\theta, d) = \sum\limits_{i=1}^{n} ||f_\theta(x_i) - y_i||^2
\end{equation}

INRs typically utilize periodic activation functions that are ideal for representing complex natural signals~\cite{sitzmann2020implicit}. These activation functions have continuous derivatives, making them well-suited for capturing continuous signals.

\subsection{INR Compression Advantages}
INR offers several advantages over other compression methods:

\begin{itemize}[leftmargin=*]
\item \textbf{Flexibility and generality.} 
INR provides exceptional flexibility, enabling effective modeling of diverse complex functions. It can handle various types of data, including images, 3D shapes, and time series. This versatility makes it suitable for tasks like image synthesis, generative modeling, and data reconstruction. Moreover, INR exhibits strong generalization capabilities, performing well on unseen or out-of-distribution data,  as the network learns meaningful representations capturing the underlying structure of the data.

\item \textbf{Storage efficiency.} 
INR compresses data into compact forms by encoding the entire function within network weights. This results in more storage-efficient representations compared to explicit representations that require individual storage for each data point. It facilitates the storage and retrieval of large-scale models and datasets.



\item \textbf{Continuity and Smoothness.} 
INR's continuity and smoothness properties are desirable for animation, interpolation, and solving partial differential equations in scientific problems.


\end{itemize}

\begin{figure*}[ht!]
\centering
\subfigure{\includegraphics[width=0.85\linewidth]{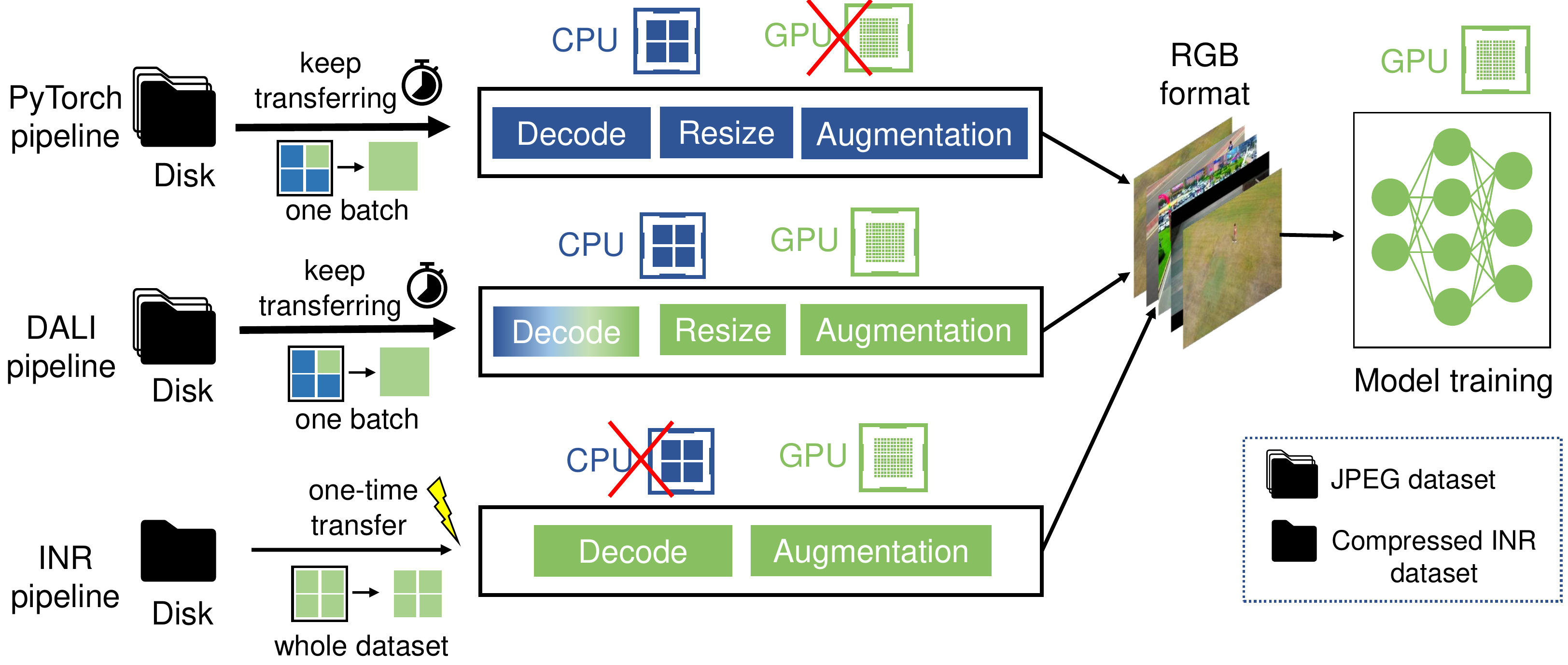}}
\caption{A high-level overview of three different training pipelines. \textbf{PyTorch pipeline:} Keep fetching batches of JPEG images from the disk to the CPU and decode to RGB format. Then do resizing and augmentation to get prepared for training. \textbf{DALI pipeline:} Similar to PyTorch pipeline. The difference is that the decoding is in hybrid mode using CPU and GPU together for acceleration. \textbf{INR pipeline:} Only transfer the whole dataset in INR MLP weights format from disk to GPU one time before training starts. Then decode the images to RGB format on-the-fly. 
}
\label{fig: three pipelines}
\end{figure*}

\subsection{INR Related Works}

We highlight two prior works that explore the utilization of INR for image compression: COIN~\cite{dupont2021coin} and COIN++~\cite{dupont2022coin}. COIN, which is the pioneering paper in this area, systematically examines the performance of INR in image compression. It achieves a higher image reconstruction quality than JPEG when the compression rate is high by encoding each image using an overfitted MLP that maps pixel locations to RGB values. COIN achieves this without employing entropy coding or learning a weight distribution, and by only quantizing the weights to 16 bits. However, it cannot achieve a better reconstruction quality than JPEG when the compression rate is small.


COIN++ builds upon COIN by incorporating meta-learning techniques to expedite INR image encoding. Additionally, it explores various quantization strategies to enhance compression. COIN++ goes further in reducing the size of network weights, resulting in a higher compression rate under the same reconstruction quality. However, it should be noted that the decoding speed of COIN++ is three times slower compared to COIN.


\subsection{Two Common Training Pipelines}

Two widely used computer vision neural network training pipelines are the PyTorch pipeline~\cite{PyTorch} and the recently introduced new training pipeline using Data Loading Library (DALI)~\cite{DALI} developed by NVIDIA, as shown in Fig.~\ref{fig: three pipelines}.

The PyTorch pipeline involves prefetching image batches from the disk to the CPU during training. The image decoding, resizing, and augmentation are performed on the CPU. While PyTorch's data loader can largely increase the speed of image preprocessing before sending the images to the GPU, it often requires multiple CPU workers to run concurrently. The overall performance of this pipeline is highly dependent on the CPU's capabilities. The presence of a weak CPU can prevent full utilization of the GPU's CUDA cores.

In contrast, DALI pipeline accelerates the JPEG decoding by leveraging both the CPU and the GPU. The resizing and data augmentation tasks are shifted to the GPU for maximizing its computational capabilities. DALI exhibits promising speedup compared to the PyTorch pipeline. However, it still heavily depends on the presence of powerful CPUs to ensure efficient data processing.



\begin{figure*}[ht!]
\centering
\subfigure{\includegraphics[width=0.86\linewidth]{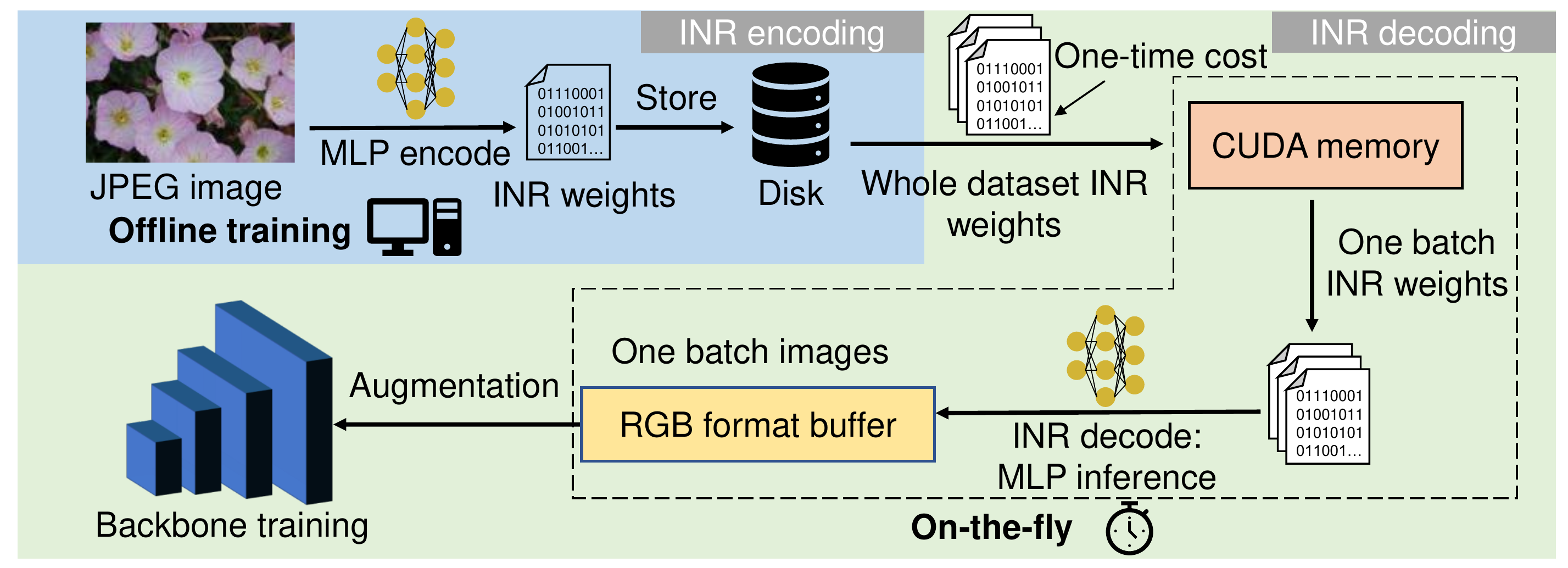}}
\caption{The encoder-decoder architecture of Rapid-INR. In the INR encoding part, each image is encoded to INR weights using a separate MLP. The encoding part is offline, and we store the whole dataset in INR weights format on the disk.  In the decoding part, the whole dataset INR weights will be transferred to CUDA memory first, and then decode one batch needed for backbone training on-the-fly when training starts. 
}
\label{fig:encoder-decoder architecture}
\end{figure*}

\subsection{Key Insights}

To the best of our knowledge, there are no existing works that have combined INR image compression and training acceleration. By leveraging the high image compression rate of INR, it becomes possible to compress the entire dataset into the memory of a single device. This enables offloading the whole training process to GPU, achieving on-the-fly decoding of images on the GPU during training, eliminating the need for repeated external memory access, data transmission, and powerful CPUs for data loading and preprocessing.

For instance, the training set of ImageNet~\cite{russakovsky2015imagenet} occupies approximately 138GB, which exceeds the on-chip memory capacity of most training devices. However, by encoding the dataset in the INR format, it only requires around 14GB, which can be directly stored in the CUDA memory. Furthermore, storing the MLP weights for INR in tensor format ensures compatibility with a wide range of GPUs and offers convenience during usage. The differences between our Rapid-INR pipeline and two previous pipelines are shown in Fig.~\ref{fig: three pipelines}.



\section{Encoder-decoder Architecture}

Rapid-INR utilizes an encoder-decoder architecture for image compression, as shown in Fig.~\ref{fig:encoder-decoder architecture}. The encoder takes images in the JPEG format and encodes them into the INR weights format, saving the encoded dataset on disk. On the other hand, the decoder decodes the images from INR weights format back to RGB format. These decoded RGB images are then used for backbone training.


\subsection{Encoder Architecture}

Rapid-INR utilizes a simple multilayer perceptron (MLP) as the INR encoder. The MLP takes a two-dimensional input representing the spatial location $(x, y)$ of each pixel in an image. The output dimension of the MLP is set to 3, representing the RGB value of a pixel. The choice of the number of layers $N$ and hidden dimensions $H$ is determined based on the image size. Larger images require more layers and hidden dimensions for optimal performance.

To determine the optimal architecture, we incorporate neural architecture search into our methodology, which will be explained in detail in Section~\ref{sec:NAS}. During training, each image in the dataset is paired with a dedicated MLP. The number of copies of INR weights matches the total number of images in the dataset.

During the training process, the MLP takes the normalized spatial location of each pixel within an image, ranging from 0 to 1, as the input. The RGB values associated with each pixel serve as the training labels. The MLP effectively learns the spatial color information, which is implicitly encoded in the MLP weights. After the offline training phase, the learned weights are transferred to disk for subsequent backbone training.


\subsection{Decoder Architecture}

The INR decoder is positioned before the backbone network to transform images from the INR weights format to the RGB format. The decoder utilizes the same MLP architecture as the encoder. Before commencing the backbone training, the entire dataset in the INR weights format is transferred from disk to the GPU's CUDA memory at one time. The compression of images into the INR format enables the accommodation of the entire dataset within the CUDA memory, avoiding external data access during training.



During backbone network training, the decoder selects a batch of images in the INR weights format from the CUDA memory and then mapped to the MLP for decoding. 
Thanks to the continuous nature of the INR's representation function, which can support an arbitrary image resolution, there is no need for image resizing during the decoding process. This implies that the images can be decoded to any desired size without loss of information or quality.

To fully harness the computational power of the CUDA cores, the decoding process is optimized to leverage pixel-level parallelism. This enables simultaneous decoding of each image within the batch, as well as parallel decoding of individual pixels within each image. GPU acceleration is also utilized for data augmentation, where the augmented images are then forwarded to the backbone for training. Minimizing external memory access is prioritized, ensuring that communication overhead is minimized. As a result, a significant portion of the computation takes place on the GPU, reducing the need for frequent data communication and optimizing overall performance.


\begin{figure*}[ht!]
\centering
\subfigure{\includegraphics[width=0.85\linewidth]{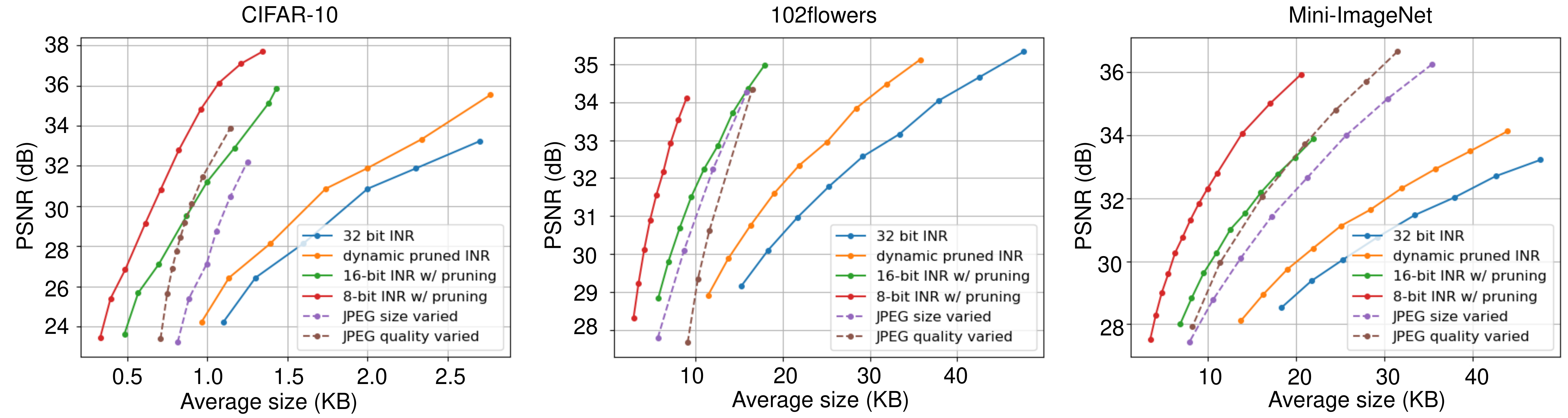}}
\caption{The relationship between PSNR and average image size using different compression techniques to INR and JPEG. (Note: The average pruning ratio for CIFAR-10 dataset is 15\%, for 102flowers and Mini-ImageNet is 25\%. 16-bit and 8-bit INR combines quantization and pruning together. 
}
\label{fig:PSNR-memory size relationship}
\end{figure*}

\begin{figure*}[ht!]
\centering
\subfigure{\includegraphics[width=0.85\linewidth]{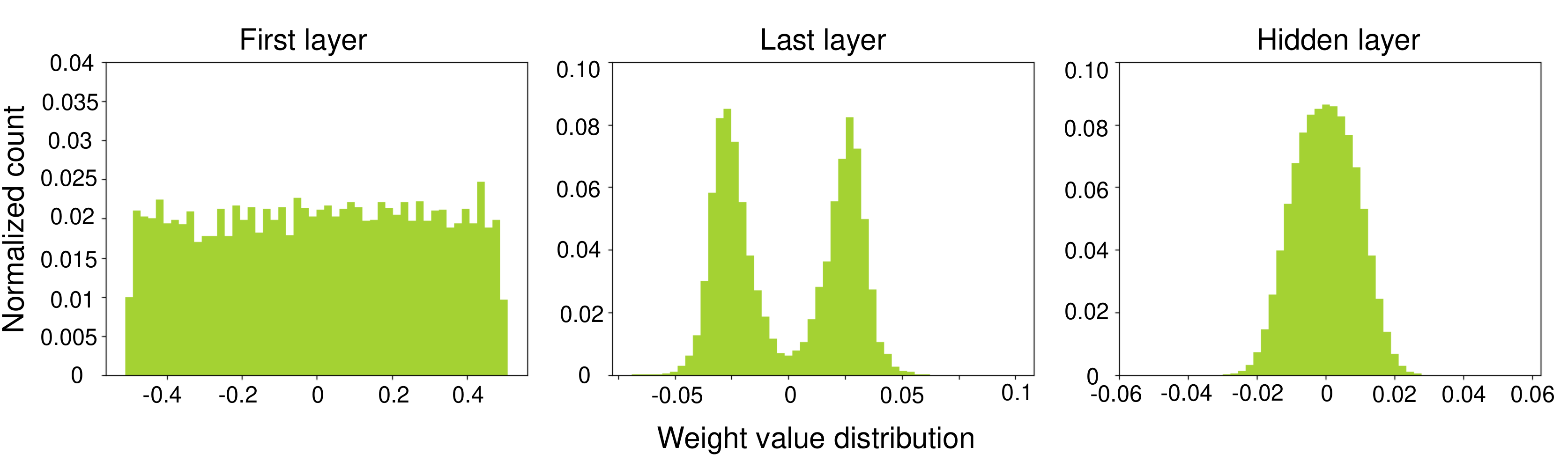}}
\caption{The normalized weight distribution of three different types of layers. First layer and last layer weight do not distribute around 0 and have a relatively sparse distribution compared with the hidden layer.  }
\label{fig:weight distribution}
\end{figure*}

\section{INR compression techniques}

Although INR can achieve similar image reconstruction quality to JPEG, it is important to consider that without employing any network compression techniques on the MLP weights for INR may result in larger file sizes compared to JPEG. To address this concern, Rapid-INR integrates two network compression methods: iterative and dynamic pruning, along with layer-wise quantization. These techniques are specifically designed to reduce the size of MLP weights for INR while minimizing the potential impact on reconstructed image quality. The goal is to strike a balance between achieving compact representations and maintaining satisfactory image reconstruction quality. We use three image datasets: Mini-ImageNet~\cite{Mini-ImageNet}, 102flowers~\cite{nilsback2008automated} and CIFAR-10~\cite{krizhevsky2009learning} with varied image sizes to evaluate our proposed techniques. We use peak signal-to-noise ratio (PSNR) as the image reconstruction quality evaluation metric. Our analysis results are shown in Fig.~\ref{fig:PSNR-memory size relationship}.

\subsection{Iterative and Dynamic Pruning}
\textbf{Insights.} Our preliminary study reveals a linear relationship between the size of the INR and the reconstructed image quality (PSNR). Additionally, different images exhibit varying reconstructed quality using the same sized INR due to color richness and variety. Moreover, our investigation indicates that low-quality reconstructed images have a larger impact on the final backbone training accuracy. Based on these findings, we propose iterative and dynamic pruning, where the pruning ratio is dynamically selected based on the PSNR of the reconstructured images. 

\textbf{Technical details.}
Our approach for Mini-ImageNet and 102flowers datasets involves two rounds of iterative pruning to achieve an optimal balance between reconstructed image quality and INR size. In the first round, we employ L1-unstructured pruning to set 20\% of the weights to 0. Subsequently, we apply the pruning mask to the weights and retrain the MLP with a smaller learning rate to help network to recover. This pruning of redundant weights facilitates network convergence, and in some cases, we even observed a slight improvement in PSNR. For the CIFAR-10 dataset with smaller image sizes, we omit the first round of pruning and directly proceed to the second round.



The second round of pruning is referred to as dynamic pruning. 
The pruning ratios $Pr$ for the CIFAR-10, 102flowers, and Mini-ImageNet datasets are defined by Eq.~\ref{eq:cifar-10},~\ref{eq:102flowers} and \ref{eq:mini-imagenet} respectively:


\begin{equation}
\label{eq:cifar-10}
Pr = 
\begin{cases}
0 & PSNR < 30\\
0.05 * PSNR - 1.5 & 30 \leq PSNR \leq 35\\
0.25 & PSNR > 35
\end{cases}
\end{equation}

\begin{equation}
\label{eq:102flowers}
Pr = 
\begin{cases}
0.2 & PSNR < 35\\
0.04 * PSNR - 1.2 & 35 \leq PSNR \leq 40\\
0.4 & PSNR > 40
\end{cases}
\end{equation}

\begin{equation}
\label{eq:mini-imagenet}
Pr = 
\begin{cases}
0.2 & PSNR < 35\\
0.04 * PSNR - 1.2 & 35 \leq PSNR \leq 40\\
0.4 & PSNR > 40
\end{cases}
\end{equation}

After the dynamic pruning, the average pruning ratios achieved on the CIFAR-10, 102flowers and Mini-ImageNet datasets are 15.6\%, 24.2\% and 25.5\% respectively.


\begin{figure*}[ht!]
\centering
\subfigure{\includegraphics[width=0.95\linewidth]{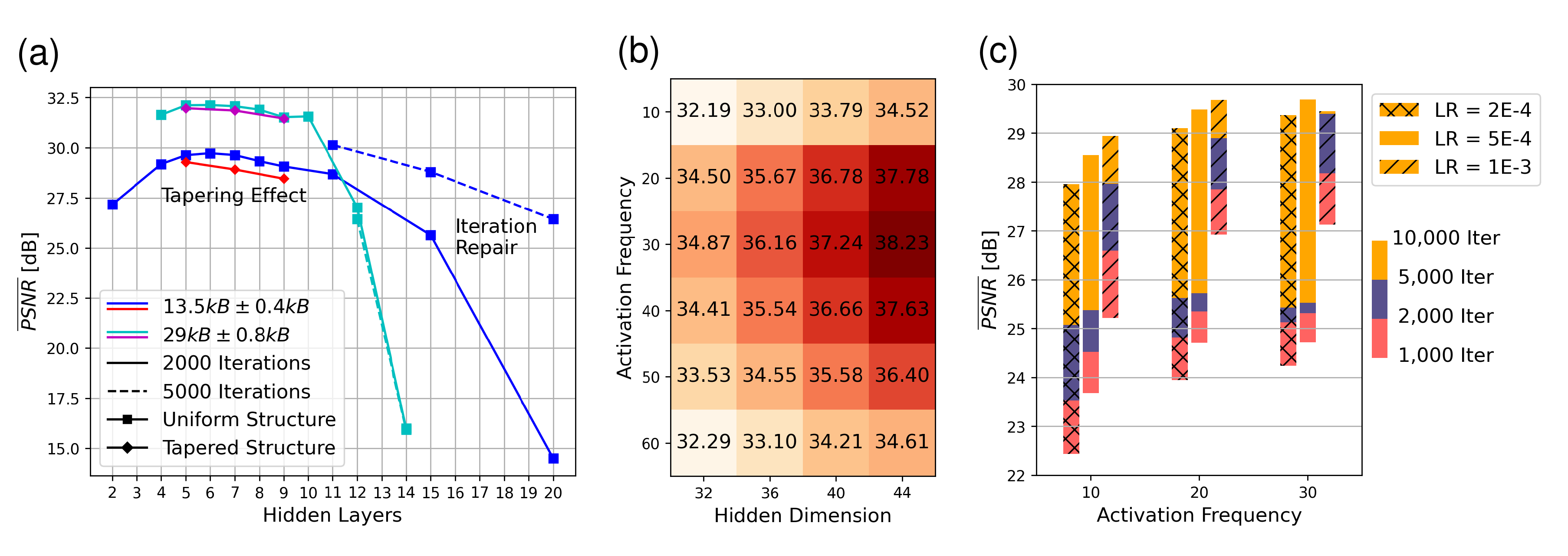}}
\caption{Design space exploration and hyper-parameter selection. (a) The relationship between the PSNR and the number of layers. It also shows the influence of different MLP architectures, different number of training iterations. (b) The relationship between the PSNR and the activation function frequency under different MLP hidden dimensions. (c) The influence of different learning rates to the PSNR when training for different numbers of iterations using different activation frequencies. 
}
\label{fig: DSE}
\end{figure*}

\subsection{Layer-wise Quantization}

\textbf{Insights.} 
We conduct a weight distribution analysis on different layers, and the results for the Mini-ImageNet dataset are depicted in Figure~\ref{fig:weight distribution}. The MLP layers exhibit distinct weight distributions after completing INR encoding training. The first layer of the MLP focuses on extracting initial features and rescaling input values. Consequently, the weights in this layer are distributed across a wide range, resembling a uniform distribution. The last layer of the MLP is responsible for regression to output normalized RGB values. Hence, the weight distribution in this layer is characterized by weights that are not concentrated around zero. The hidden layers of the MLP are involved in feature extraction and learning. The weight distribution in these layers follows a Gaussian distribution, with the majority of weights centered around zero.


The varied distribution patterns of different types of weights are not suitable for the same quantization strategy, and also weights with a Gaussian distribution within a small scale are more suitable for quantization. To reduce the performance degradation caused by quantization, we adopt a layer-wise quantization approach, quantizing only the weights of the hidden layers. This preserves the first and last layer weights in full precision, mitigating the performance degradation effectively.


\textbf{Technical details.}
In our layer-wise quantization, only the weights of the hidden layers are quantized to 8 bits while retaining the weights in the first and last layers in their full precision of 32 bits. By implementing layer-wise quantization, we can effectively prevent a substantial drop in the PNSR and maintain storage efficiency.

\section{Neural Architecture Search and Hyperparameter Tuning}
\label{sec:NAS}

To optimize the reconstruction quality of INR-compressed images, the selection of the MLP architecture and hyperparameters plays a crucial role. In order to find the optimal MLP architecture within the given storage constraints and determine the best hyperparameters such as learning rate and activation function frequency, we conduct a neural architecture search and hyperparameter tuning. These processes involve systematically exploring different architectural configurations and hyperparameter settings to identify the combination that yields the highest performance. By employing these techniques, we were able to improve the encoding capabilities of INR.


\subsection{Neural Architecture Search}

We conduct experiments with two different sizes of INR: 13.5KB and 29KB. To explore the effects of different MLP architectures, we vary the number of hidden layers and hidden dimensions in tandem, while maintaining a fixed size constraint. We further consider two types of architectures: uniform and tapered. The uniform architecture has the same dimension for all hidden layers, while the tapered architecture gradually increases and then decreases the dimensions of consecutive hidden layers.

Based on our findings, as shown in Fig.~\ref{fig: DSE} (a), the uniform architecture outperforms the tapered architecture in terms of PSNR. We also use different training iterations for these two MLP architectures. Increasing the training iterations improves the PSNR for the 13.5KB INR variant. We also discover that excessively deep MLP architectures (with more than 10 layers) result in decreased PSNR. Therefore, it is recommended to limit the number of layers to less than 10 for optimal performance.




\subsection{Hyperparameter Tuning}

We also investigate the impact of two key parameters: the activation function frequency and the learning rate used during INR encoding training. Fig.~\ref{fig: DSE} (b) presents the PSNR results obtained by varying the activation function frequency and hidden layer dimensions. Our findings indicate that an activation function frequency of 30 yields the best performance in terms of PSNR.

Furthermore, we examine the influence of different learning rates in conjunction with varying numbers of training iterations. The results are depicted in Fig.~\ref{fig: DSE} (c). Starting at a learning rate of 1e-3 and above, image compression fails for a subset of the images. From these observations, it is suggested to select a learning rate within the range of 2e-4 to 5e-4 to achieve optimal performance.



\section{Experiment Results}

\subsection{Experiment Setup}


Our experiment is conducted using PyTorch with an NVIDIA A6000 GPU and an Intel 6226R CPU. The architecture of the INR MLP varys for each dataset: 3 layers with 15 hidden dimensions for CIFAR-10, 10 layers with 32 hidden dimensions for 102flowers, and 10 layers with 40 hidden dimensions for Mini-ImageNet. The MLP employs the sine function as the activation function with a frequency value of 30. The training of the INR MLP consists of three rounds. In the first round, we train the MLP to overfit using full precision. The learning rate for this stage was 5e-4, and we train for 5,000 epochs. In the second and third rounds, we apply iterative and dynamic pruning techniques. The learning rate for these rounds is set to 2e-4, and we train for 10,000 epochs. To optimize gradient descent, we utilize the Adam optimizer and incorporated cosine learning rate decay. This technique progressively decreases the learning rate as the gradient diminished during the training process.



For the image classification task, we select ResNet-18 as the backbone network. It is important to note that this choice is specific to our case study, as Rapid-INR can be easily extended to other vision tasks and backbone architectures. During the backbone training, we employ a batch size of 128 for CIFAR-10 and 64 for both 102flowers and Mini-ImageNet datasets. The learning rate used for backbone training is set to 1e-2, with a moment decay of 5e-4. We utilize the SGD optimizer for gradient descent and train for 200 epochs.

\begin{figure}[h!]
\centering
\subfigure{\includegraphics[width=1.00\linewidth]{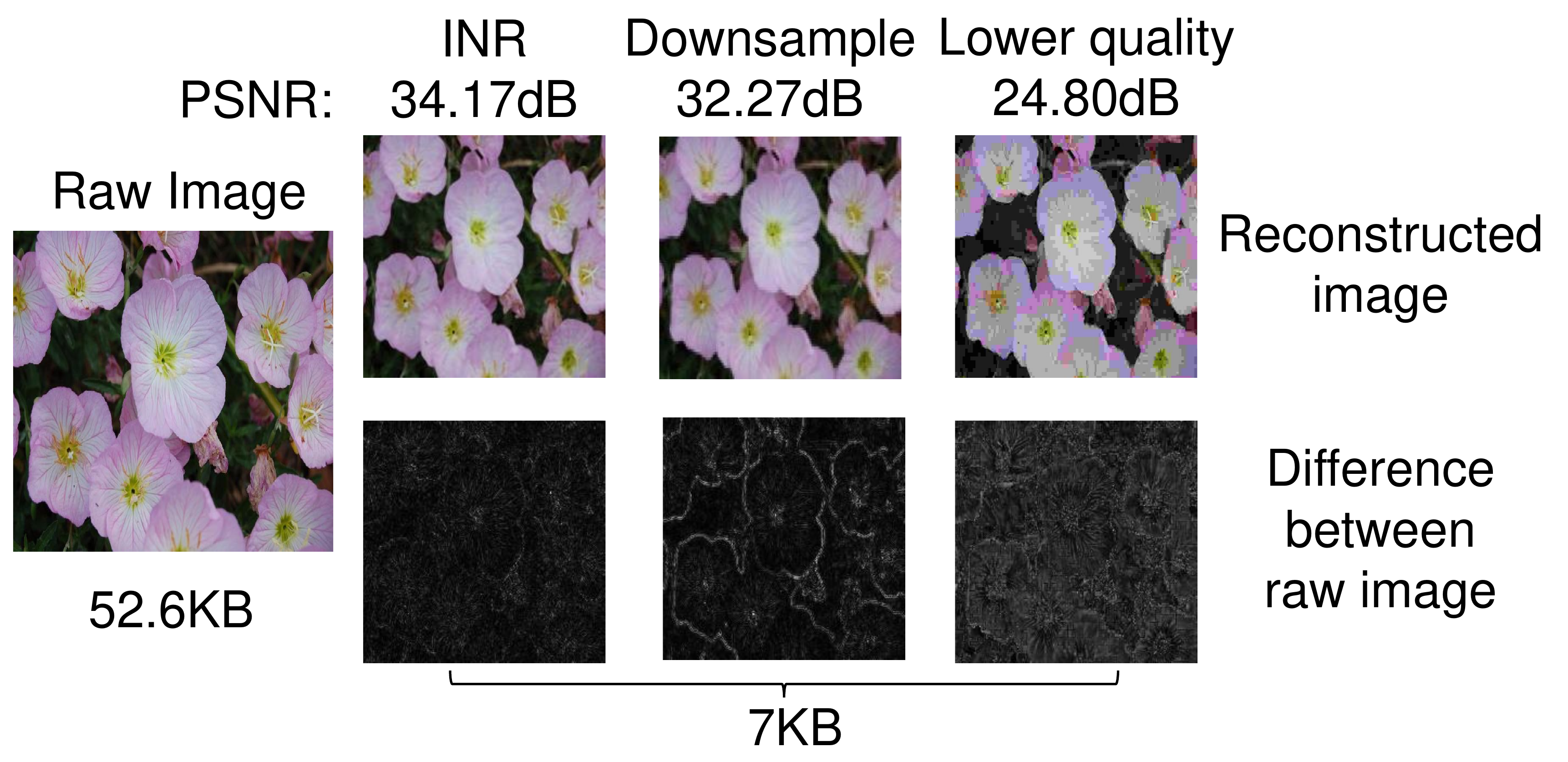}}
\caption{Image compression techniques comparison. INR can provide a higher reconstruction quality under the same image size. All three of these techniques struggle to handle object boundaries effectively. 
}
\label{fig:image reconstruction}
\end{figure}

\begin{figure}[ht!]
\centering
\subfigure{\includegraphics[width=0.88\linewidth]{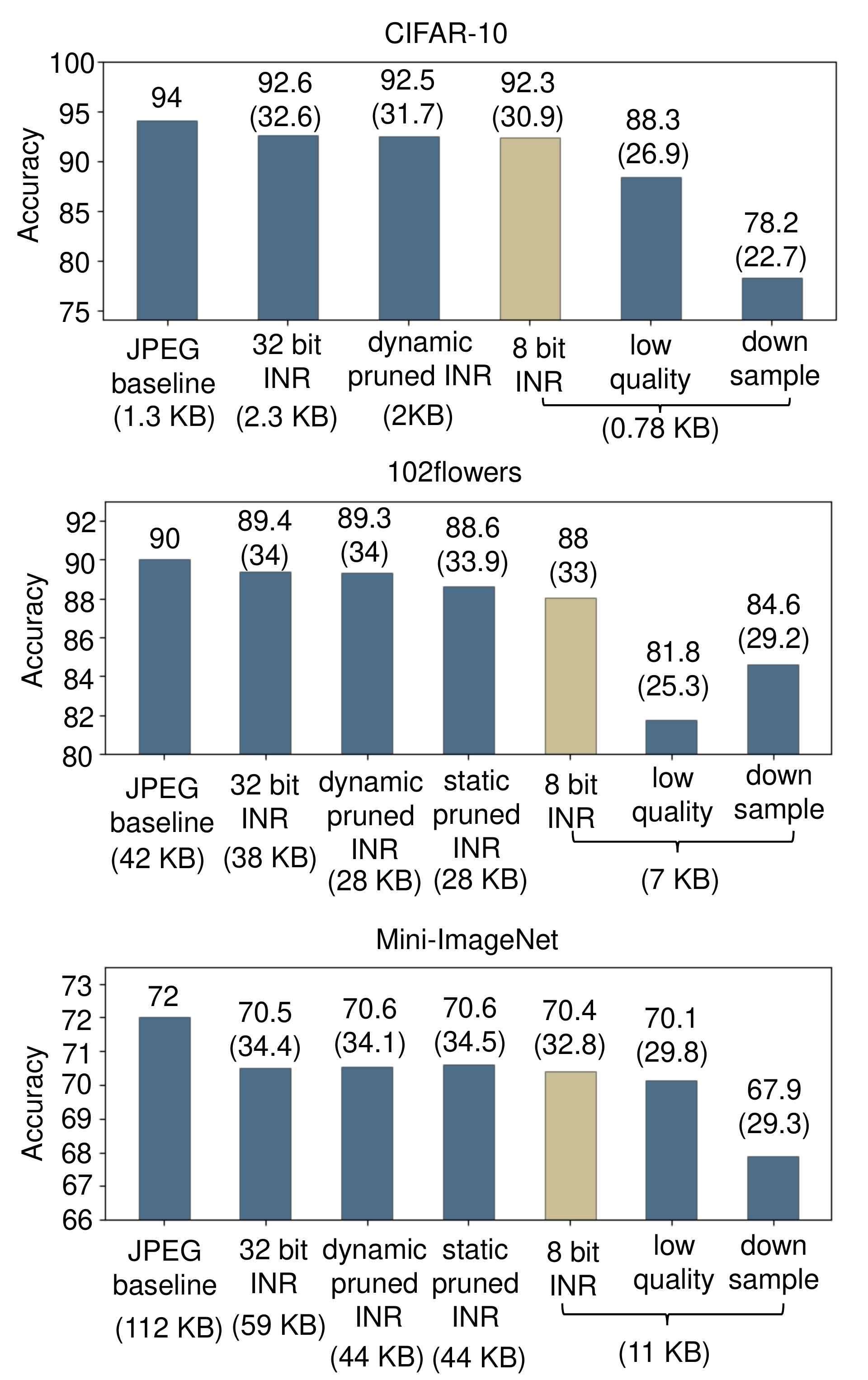}}
\caption{Backbone training accuracy comparison using INR encoded images and JPEG images. The accuracy (upper number) and corresponding PSNR (lower number) is annotated above the bar. INR-encoded images only cause a slight backbone training accuracy loss. Compared with JPEG images in the same average size, INR encoded images can provide a higher backbone training accuracy.}
\label{fig: Backbone training accuracy}
\end{figure}

\begin{figure*}[ht!]
\centering
\subfigure{\includegraphics[width=0.86\linewidth]{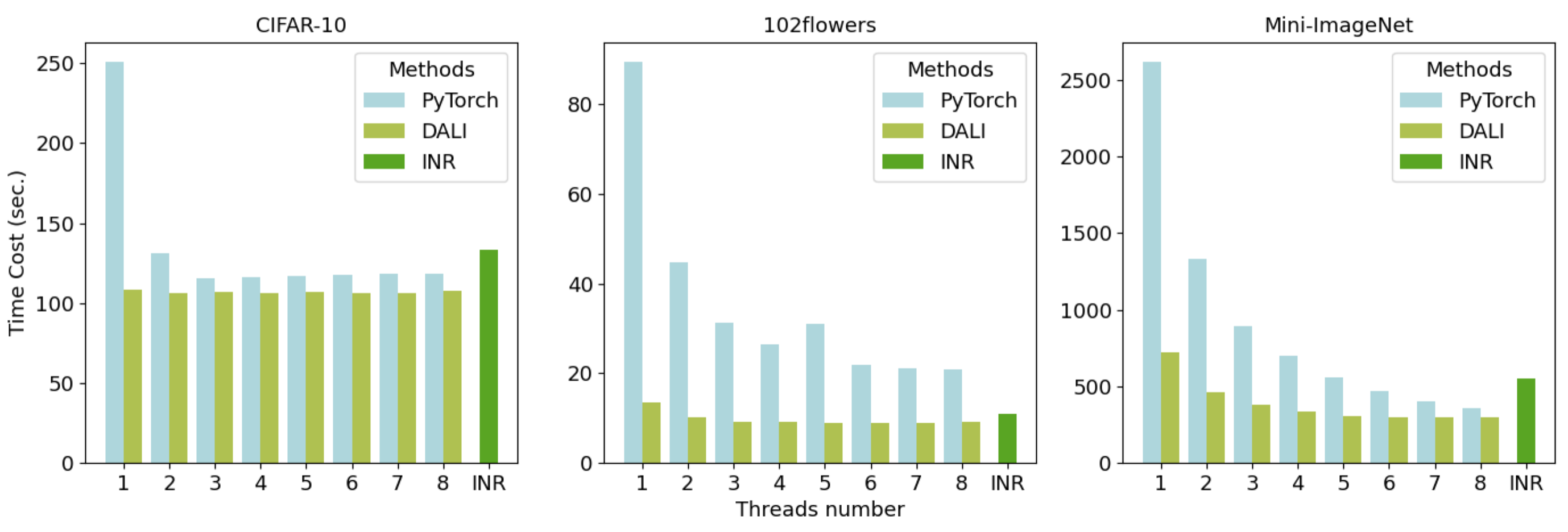}}
\caption{Backbone training time comparison across different pipelines and datasets. "Thread number" refers to the number of CPU threads utilized for image pre-processing. The reported total time cost represents the end-to-end training time for 10 epochs.}
\label{fig: backbone training time comparison}
\end{figure*}

\subsection{Reconstruction Quality}

In our experiment, we compare three different image compression techniques: INR, JPEG with different qualities, and JPEG downsampling. Fig.~\ref{fig:image reconstruction} showcases the results of reconstructing an image from the 102flowers dataset using these methods. The images presented include the original image, the reconstructed image using each technique, and the corresponding residual image.

When considering the same memory space, the INR-reconstructed image exhibits the highest quality among the three techniques. However, all of them share a common issue: the sharp boundaries in the images have been blurred. This blurring effect is more evident in the residual image, where the boundary of the flower appears clearer compared to other regions.

The blurring effect in the INR-reconstructed image can be attributed to the nature of the learned function in INR, which is continuous. Although INR can capture some useful information about the boundary, it struggles to accurately model the abrupt color changes that occur at boundaries. On the other hand, JPEG compression involves approximation techniques, and since the boundary region occupies only a small portion of the image, the color information in this region is often blurred due to the influence of surrounding regions during compression.


\subsection{Backbone Training Accuracy}

We further analyze the accuracy of training the backbone network by utilizing images decoded from various compression techniques. The experimental results are presented in Fig.~\ref{fig: Backbone training accuracy}. We compare the accuracy of training the backbone using images decoded from different types of INR-encoded images, as well as raw JPEG images. Although there is only a slight drop in accuracy, when combining dynamic pruning and 8-bit layer-wise quantization, a significant amount of storage space can be saved. Moreover, we ensure a fair comparison by fixing the size of the compressed INR and JPEG images. Notably, INR with dynamic pruning and 8-bit layer-wise quantization outperforms both JPEG compression techniques in terms of backbone training accuracy.


\subsection{Hardware Speedup}
We also conduct an end-to-end training time profiling to demonstrate the accelerated performance of the INR training pipeline. It compares INR with two robust baselines (Fig.~\ref{fig: three pipelines}) across three datasets. The measured training time includes data transfer costs, decoding, augmentation, and neural network training.



The PyTorch and DALI pipelines share the same training flow and configuration as the INR pipeline, but differ significantly in the data preparation phase. Firstly, the baselines require frequent data transfers between CPU and GPU memory, while INR only requires a single transmission due to its superior compression rate. Secondly, the baselines heavily rely on CPU power for training since GPU memory cannot store JPEG format data. This means the CPU must decode small JPEG images into large tensor format, increasing the CPU's transmission workload. Furthermore, the decoding algorithm itself can become a bottleneck due to limited parallelization. Thirdly, the baselines struggle to fully utilize GPU resources, including GPU memory and parallel computation, unlike the INR pipeline. INR achieves higher GPU memory utilization by accommodating all training data within the GPU and keeping it fixed. Additionally, INR's decoding process leverages on-the-fly GPU parallel computation instead of sequential CPU processing.


Fig.~\ref{fig: backbone training time comparison} presents profiling results from three datasets, showcasing the scalability of INR across different scenarios. The datasets have varying characteristics: CIFAR-10 has small-sized images but a large dataset size, 102flowers has large-sized images but a small dataset size, and Mini-ImageNet contains large-sized images and a substantial dataset size. (Note that we use Mini-ImageNet for profiling convenience, but the original ImageNet dataset can also fit into the NVIDIA A6000 card.) In the experiment, we increase the number of threads to accelerate the pipeline until it reaches the hardware limits. On 102flowers and Mini-ImageNet datasets, INR outperforms the PyTorch pipeline with eight threads and four threads, respectively, and both outperform DALI with one thread. However, on CIFAR-10, INR only surpasses the PyTorch pipeline when using one thread. The key distinction between the 102flowers/Mini-ImageNet datasets and CIFAR-10 is the image size. Larger image sizes exploit parallelism more effectively, which explains the decreasing INR training time with increasing CIFAR-10 batch sizes. However, using large batch sizes may compromise training accuracy despite achieving speedup.

\section{Conclusions}

In this paper, we propose \textbf{Rapid-INR}, an INR-based image compression technique that effectively reduces image size while preserving acceptable quality. Rapid-INR enables accelerated training in computer vision tasks by compressing the training dataset, storing it on GPU, and performing on-the-fly decoding. We also introduce dynamic pruning and layer-wise quantization techniques to achieve further compression with minimal quality loss, building upon previous research.

While Rapid-INR demonstrates improved storage efficiency compared to the JPEG image format and achieves speedup in image classification training compared to PyTorch and DALI pipelines, it still faces several challenges, like the lack of interpretability, computational complexity, memory efficiency in INR encoding, and sensitivity to hyperparameters. These factors need to be carefully evaluated to determine the suitability of INR for specific tasks or applications.

\section{Acknowledgements}
This work and its authors are partially supported by the National Science Foundation under Grant No.2202329, Cisco, and Samsung. The authors would also like to thank Xuebin Yao and Pradeep Subedi from Samsung for their insightful discussions.

\bibliographystyle{IEEEtran}
\bibliography{reference}

\end{document}